\documentclass[letterpaper, 10 pt, conference]{ieeeconf}  
\IEEEoverridecommandlockouts                              
\usepackage{graphics} 
\usepackage{epsfig} 
\usepackage{algorithm}
\usepackage{algorithmic}
\usepackage{verbatim}
\usepackage{cite}

\title{\LARGE \bf Toward Automated Programming for Robotic Assembly Using ChatGPT}

\author{Annabella Macaluso$^{1}$, Nicholas Cote$^{2}$, and Sachin Chitta$^{2}$
\thanks{$^{1}$Annabella Macaluso is with the University of California San Diego, La Jolla, CA 92093, USA
        {\tt\small amacalus@ucsd.edu}}%
\thanks{$^{2}$Nicholas Cote and Sachin Chitta are with Autodesk Research (Robotics), Autodesk Inc.
        {\tt\small nick.cote@autodesk.com}}%
}

\begin{document}
\maketitle
\thispagestyle{empty}
\pagestyle{empty}

\begin{abstract}
Despite significant technological advancements, the process of programming robots for adaptive assembly remains labor-intensive, demanding expertise in multiple domains and often resulting in task-specific, inflexible code. This work explores the potential of Large Language Models (LLMs), like ChatGPT, to automate this process, leveraging their ability to understand natural language instructions, generalize examples to new tasks, and write code. In this paper, we suggest how these abilities can be harnessed and applied to real-world challenges in the manufacturing industry. We present a novel system that uses ChatGPT to automate the process of programming robots for adaptive assembly by decomposing complex tasks into simpler subtasks, generating robot control code, executing the code in a simulated workcell, and debugging syntax and control errors, such as collisions. We outline the architecture of this system and strategies for task decomposition and code generation. Finally, we demonstrate how our system can autonomously program robots for various assembly tasks in a real-world project.
\end{abstract}

\section{INTRODUCTION}

The way robots are programmed for adaptive assembly has progressed significantly over the years. Initially, robots were programmed manually, either through a teach pendant or by guiding the robot physically through desired motions. Offline Programming (OLP) software later enabled robots to be programmed, simulated, and optimized on a computer and Parametric Design workflows further streamlined the process of extracting tool-paths and targets from CAD (Computer Aided Design) geometry. Today, Machine Learning enables robots to adapt to variability in a design or workcell, significantly reducing the need for task-specific programming.

Despite these advancements, the process of programming, testing, and debugging such systems is labor-intensive, time-consuming, and involves a lot of trial and error, resulting in highly specialized code tailored for a specific product. Moreover, it requires deep expertise in multiple domains, such as robotics, perception, manufacturing, and software engineering, which poses a barrier to adoption in industry. While this approach may be suitable for low-mix, high-volume manufacturing, it lacks the flexibility needed to adapt to a diverse range of assembly tasks, highlighting the need for a more general approach. 

One wonders: \textit{Is it possible to automate this process?}

\begin{figure}[thpb]
    \centering
    \includegraphics[scale=1.0, width=3.2in]{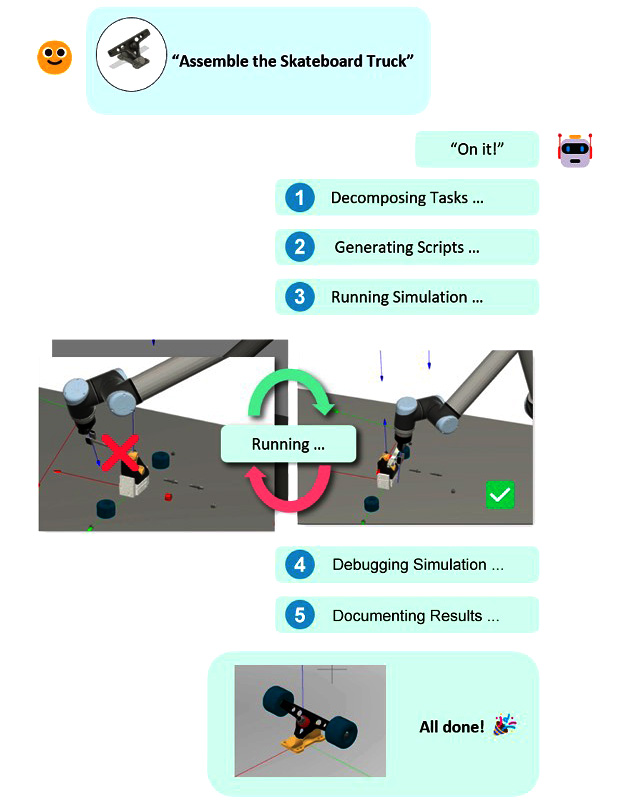}
    \caption{Our workflow utilizes GPT-4's generalization and code-writing abilities to contextualize robotic workcell and CAD information in order to generate code for assembly tasks such as "Assemble the Skateboard Truck".}
    \label{fig:front_page}
\end{figure}

Recent developments in Large Language Models (LLM) \cite{gpt-4, llama, minigpt-4}, like ChatGPT, have shown great promise in answering this question. Specifically, LLMs have shown the capacity to understand and process natural language instructions, the ability to generalize from examples to new tasks, and the ability to write code. We believe these capabilities can be harnessed and applied to real-world challenges in the manufacturing industry and, furthermore, may represent an opportunity to shift the burden of developing adaptive robotic assembly systems from people to LLMs.

In this paper, we present a novel workflow that uses ChatGPT to automate the process of programming robots for adaptive assembly by decomposing complex tasks into simpler subtasks, generating robot control code, executing the code in a simulated workcell, and debugging syntax and control errors, such as collisions. We outline the architecture of our workflow and strategies we employed for task decomposition and code generation. Finally, we demonstrate how our system can autonomously program robots for various assembly tasks in a simulated real-world project.

\section{RELATED WORK}

The manufacturing and construction industries are transitioning from traditional methods to digital, computationally-driven \textit{design-robotics} workflows. This shift is fueled by the rise of increasingly digital workflows \cite{fabrication-1, fabrication-2} that seamlessly integrate computational design methodologies with modern robotic fabrication systems  \cite{fabrication-1, fabrication-3}. Integral to these workflows are two-way feedback loops, wherein design goals and manufacturing constraints inform one another. Real-time feedback mechanisms and automated problem solving strategies during the fabrication process further optimize this process and make it adaptive \cite{fabrication-4}. This trend leans towards methods that are driven primarily by design data, integrated with CAD software, and which significantly reduce coding and development time \cite{fabrication-5, fabrication-6}. A notable evolution in this area is the incorporation of LLMs into computational modeling and manufacturing \cite{design}, laying the groundwork for our research. 

LLMs are already making strides in robotics, as in \cite{c2, voxposer, Palm-E}. In these studies, LLMs serve as language interfaces for real-world robotic applications and scenarios. Studies by \cite{hugging-gpt, toolformer, mrkl} specifically explore tool usage with LLMs. Karpas \textit{et al.} further suggests that integrating LLMs into a system of expert modules and granting them access to external tools to help them solve specific tasks and address their inherent limitations. While GPT-4 \cite{gpt-4} is designed to handle multi-modal inputs, its public usage is limited to text-based modalities, thus, overcoming its perceptual, mathematical and task-specific constraints requires a suite of robotics tools. In \cite{c7}, Koga \textit{et al.} introduced a CAD to assembly pipeline that provides scripting tools and such a suite of high-level assembly behaviors for designers to plan and automate robotic assembly tasks. This pipeline, enriched by a task-level API, offers a toolkit that code-writing LLMs can utilize. 

Despite their challenges, LLMs offer significant promise due to their ability to process natural language, write code, and generalize across diverse tasks. Their proficiency in pattern matching for both text and numeric data without extra fine-tuning makes them even more powerful \cite{c1}. Many researchers have demonstrated this ability for robotic applications using task and workcell representations \cite{c4}. Our work leans into these strengths in order to decompose complex assembly tasks recursively into manageable subtasks and assembly behavior labels \cite{c2} and to write robotic assembly code based on the result. With these advancements in mind, the need for designers or engineers to develop application-level code for manufacturing and construction processes might soon be redundant, with LLMs poised to take on this role.

\section{ARCHITECTURE}

At a high-level, we introduce a multi-agent system that utilizes ChatGPT to generate and test Python scripts for the robotic assembly of an arbitrary design. The term \textit{agent} in this work refers to a Python class that connects to the OpenAI API, ensures secure interaction with ChatGPT, and stores and maintains the chat history. Agents are herein sub-classed and configured to solve specific problems later on. As others have remarked \cite{gpt-survey, gpt-4}, we found that GPT-4 provides better responses than other models and solely use it in this work. The default tuning parameters were also employed. We develop two specialized Agents for this workflow for task decomposition and script generation, discussed in detail later on.

The chat history shows ChatGPT agents on what is expected in the response. The entire history is provided to ChatGPT with each prompt, thus, we bootstrap the agent history with contextual information prior to submitting an initial prompt. We also group entries as follows: \textit{system guidelines}, which includes the role the agent is expected to play and rules regarding response content and formatting; \textit{task context}, which includes the design, workcell constants, reference docs, and examples; and \textit{run-time history}, which includes responses generated by ChatGPT and feedback provided from simulation. The run-time history grows throughout a session, allowing an agent to iterate and improve upon prior responses. For privacy reasons, certain terms are swapped with a corresponding public or private alias before or after an interaction with the OpenAI API.

\subsection{CAD to ChatGPT}

Although ChatGPT appears to understand natural language assembly procedures and spatial relationships for \textit{common} objects and assemblies, it's unequipped to handle 3D geometries and standard CAD representations (e.g. STLs). While it's indeed possible to convey some geometrical information to ChatGPT, we observed that presenting a dictionary of \textit{assembly information} is more useful for code generation. This information is commonly stored by default in the CAD representation of a given assembly and includes individual part names, classes, physical properties, and design poses as well meta-information such as part adjacencies, joints, sub-assemblies, and shared origin frames. A subset of this data is, then, extracted from the CAD model and saved to a JSON file and provided as text to ChatGPT downstream. To ensure that parts with technical names (i.e. manufacturer-specific serial numbers) are more readable to ChatGPT, we also annotate this file with a brief, General Language Description of each part.

\subsection{Algorithm}

Given a textual representation of a design, the following process generates a set of error-free, simulation-tested Python scripts that can be used for robotic assembly. Note that the algorithm shown doesn't include stop conditions based on the number of failed script generation attempts, errors caused by prior scripts, connection errors with OpenAI API, and so on:

Initialize a separate thread for the workcell simulation; note that a reference to the workcell will be required later on when executing Python modules. Next, initialize the Task Decomposition Agent (TDA). Presented with the design representation, it infers the assembly process decomposes it into a sequence of assembly subtasks with corresponding behavior labels. The main thread then enters a loop, continuously checking if all subtasks are completed. Once all subtasks are marked complete, the simulation thread is stopped and the main process ends. 

\textit{For each iteration of the main loop:} The next subtask, its corresponding behavior label, and any errors from prior iterations are acquired. For the acquired subtask, a dedicated Script Generation Agent (SGA) is initialized using the given behavior label. The SGA then enters an inner loop, continuously \textit{trying} to generate a successful script for the subtask. Whenever an error is caught, this loop continues and the SGA tries to generate a better script. 

\textit{For each iteration of the SGA inner loop:} The SGA generates a Python script string for the specific subtask, behavior label, and error (if present). The string is then saved locally as a Python module, allowing it to be accessed later. The Python module is imported and, if successful, checked for syntax and formatting errors. Then, the module's \verb|main| function can be called with a reference to the simulated workcell. If the module returns, the subtask is marked as done.

\begin{algorithm}
\caption{Generate Robotic Assembly Scripts}
\begin{algorithmic}[1]
\REQUIRE Textual representation of a design, $D$
\ENSURE Set of tested Python scripts for robotic assembly
\STATE $simulation \gets \texttt{WorkcellSimulation}()$
\STATE $simulation.\texttt{Start}()$  \COMMENT{separate thread}
\STATE $TDA \gets \texttt{TaskDecompositionAgent()}$
\STATE $subtasks \gets TDA.\texttt{Decompose}(D)$
\WHILE{!$subtasks.$\texttt{AllDone}}
    \STATE $subtask, label, error \gets tasks.\texttt{GetNext}()$
    \STATE $SGA \gets \texttt{ScriptGenerationAgent}(label)$
    \WHILE{$true$}
        \STATE $script \gets SGA.\texttt{Write}(subtask, error)$
        \STATE $fp \gets SGA.\texttt{Save}(script)$
        \STATE \textbf{try:}
            \STATE \hspace{0.27cm} $module \gets \texttt{ImportAndCheckModule}(fp)$
            \STATE \hspace{0.27cm} $module.\texttt{main}(simulation.\texttt{workcell})$
            \STATE \hspace{0.27cm} $subtasks.\texttt{MarkDone}(subtask)$
            \STATE \hspace{0.27cm} \textbf{break}  \COMMENT{script ran without error}
        \STATE \textbf{except} \texttt{Exception} as $e$\textbf{:}
            \STATE \hspace{0.27cm} $error \gets e$
    \ENDWHILE
\ENDWHILE
\STATE $simulation.\texttt{Stop}()$
\end{algorithmic}
\end{algorithm}

\subsection{Task Decomposition Agent}

The TDA leverages the pattern matching and generalization capabilities of LLMs to break down complex assembly tasks into a sequence of simple subtasks, as opposed to ones requiring detailed or nuanced implementation, and then assign behavior labels to them. This allows subtasks and behavior label pairs to be addressed individually during script generation.

\begin{figure}
    \centering
    \includegraphics{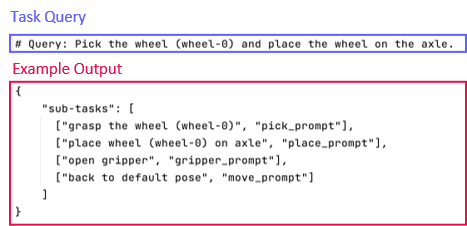}
    \caption{Example of demonstration provided for few-shot prompting. Provides context on how language output from ChatGPT should be formatted and what a "successful" example looks like. Formatting inspiration taken from VoxPoser \cite{voxposer} and Code as Policies \cite{c2}}
    \label{fig:enter-label}
\end{figure}

For each assembly subtask the TDA assigns labels based on common robotic assembly behaviors, such as \textit{Move}, \textit{Pick}, \textit{Place}, and \textit{Insert}. A task like \textit{Assemble Axle} might be decomposed into \textit{Detect Axle}, \textit{Pick-Up Axle}, \textit{Move Axle}, and \textit{Insert Axle}. For each behavior, we supply a high-quality set of demonstrations in the form of Python scripts for few-shot prompting of the SGA later on. Steering the LLMs with few-shot learning allows us to ensure a higher success rate, improve accuracy for down-stream tasks, and set an appropriate level of detail for decomposed subtasks. While we assume the topological order of subtasks is generated correctly, this is not necessarily the case, highlighting the need for an additional verification stage in future work.

We formulate the initial chat context as ($R$, $L$, $S$, $P$, $B$, $E$), where $R$ is agent role, $L$ are the formatting rules, $S$ is the assembly sequence as a dictionary, $P$ is a list of part names, $B$ is a list of available behavior primitives, and $E$ is a set of high quality examples for few-shot prompting. The user then provides a language description of the task, $T$, (e.g. "Assemble the toy car").

For a simple assembly with 10 parts, the TDA may identify as many as 40-50 subtasks, requiring the use of equally many SGAs. With multiple SGAs working in parallel, this process takes \textit{only a few minutes}.

\begin{figure}
    \centering
    \includegraphics{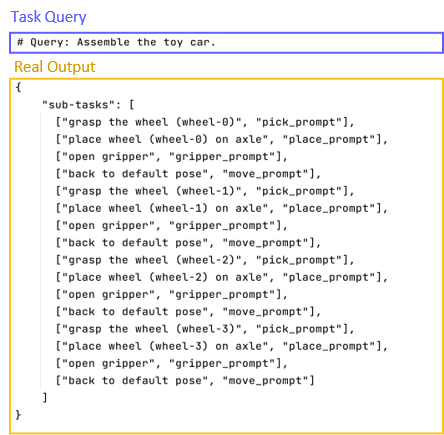}
    \caption{Example TDA input/output. User inputs a task query and the output is a structured list of subtasks each with an assigned robotic behavior primitive. In addition to the task query, information about the assembly, objects that reside in it, grippers, robots etc. are also provided as context during input.}
    \label{fig:TDA}
\end{figure}

\subsection{Script Generation Agent}

The SGA leverages the code-writing, debugging, and text formatting capabilities of LLMs to generate and debug Python scripts for a given subtask and behavior label. To do this, the user provides the subtask and behavior primitive provided by the TDA, e.g. $["Pick the baseplate", "Pick"]$. This agent leverages the capabilities of ChatGPT to write Python code and debug it, to format text, and to generalize from example code to generate solutions for specific tasks. 

We formulate the initial chat context as ($R$, $L$, $A$, $W$, $P$, $E$), where $R$ is agent role, $L$ are the scripting rules, $A$ is the assembly context as a dictionary, $W$ is the workcell context as a dictionary, $D$ is reference documentation for the Python API, and $E$ is a user-defined scripting example associated with the provided behavior primitive. The generated script, $S$, and any syntax errors or runtime exceptions, $X$, are appended to this context every iteration of the SGA, allowing the agent to improve upon prior versions: ($R$, $L$, $A$, $W$, $D$, $E$, $S_1$, $X_1$, ..., $S_n$).

Because each assembly behavior is purposefully succinct, examples often contain only a few lines of code. To increase the diversity of generated code, we provide a few, varied examples with different levels of complexity. We observed that agents were more likely to provide a correctly formatted response when formatting is demonstrated in examples, reinforced in an agent's natural language rules, and when prompts appear code-like. We also observed an ability to deduce the behaviors of functions and classes with commonly accepted naming conventions, especially when supplemented with function metadata (e.g. type-hints and docstrings) and detailed error explanations.  

It's important to note that successfully running scripts at the syntax level in a simulated environment doesn't guarantee its semantic correctness. Occasionally, we observed "debugged" motion commands enclosed in a try-except block, which executed without error but failed to complete the subtask, in which case we had to adjust either the prompt or script to achieve success.

\begin{figure}[thpb]
    \centering
    \includegraphics[scale=1.0, width=3.1in]{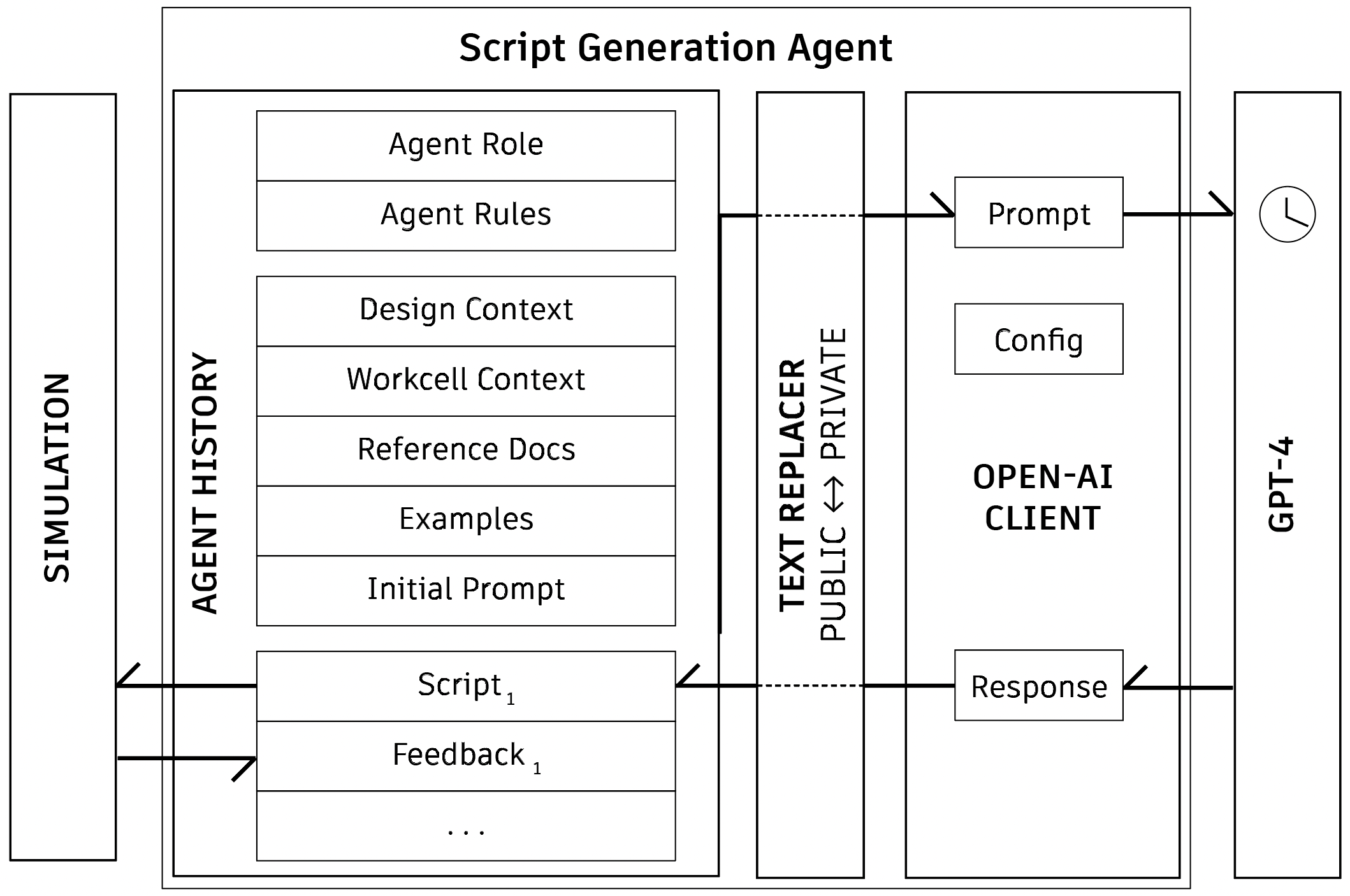}
    \caption{Script Generation Agent class architecture. The agent history gets passed to a client that communicates with ChatGPT to generate a script. The script is added to the history and executed in simulation. Then feedback is added to the history and the process repeats.}
    \label{fig:enter-label}
\end{figure}

\section{EXPERIMENTS}

\begin{figure}
    \centering
    \includegraphics[scale=1.0, width=3.1in]{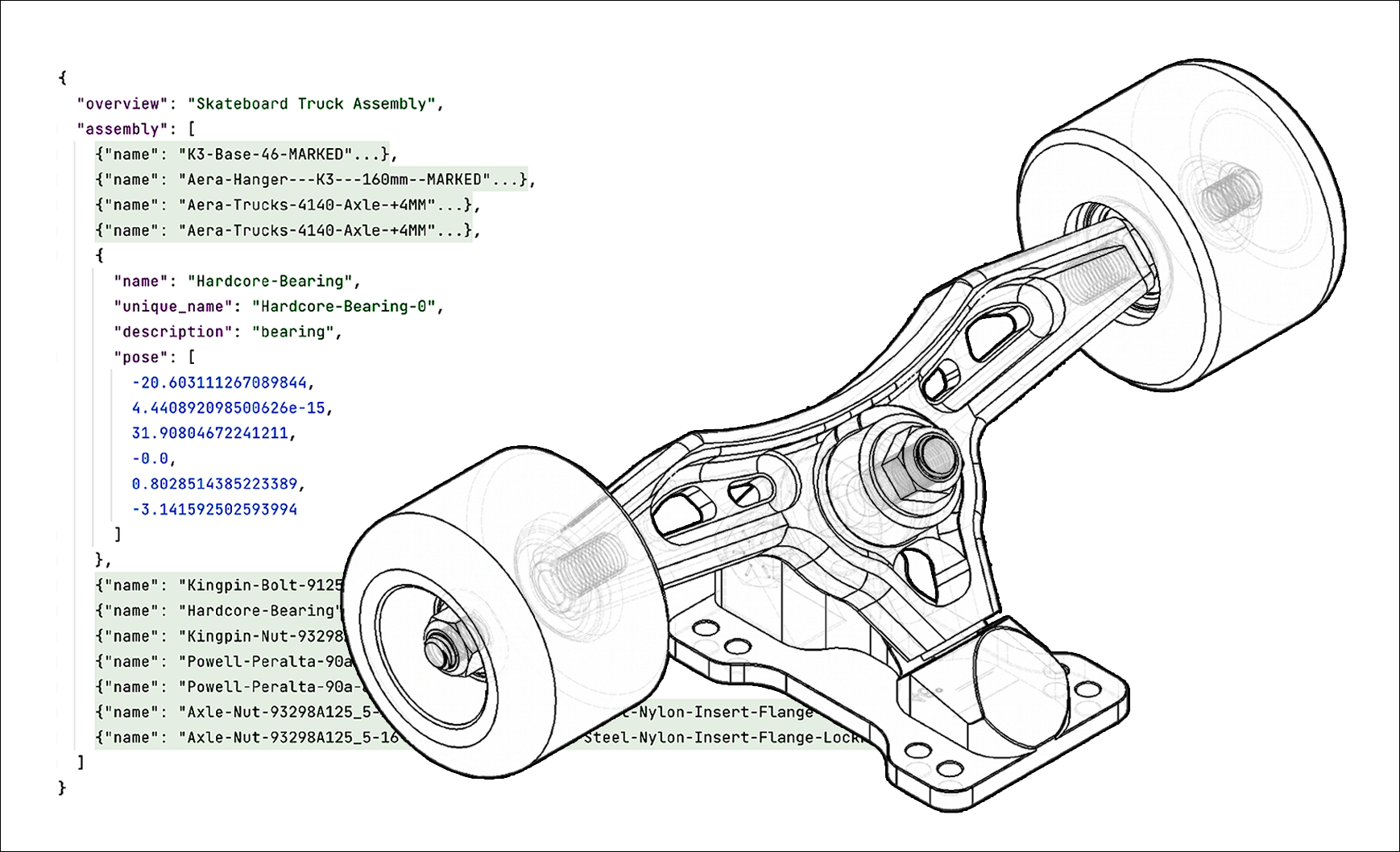}
    \caption{Geometry and assembly representations of the skateboard truck.}
    \label{fig:skateboard-truck}
\end{figure}

Here we use our workflow to assemble a \textit{Skateboard Truck} consisting of the following parts: \textit{Kingpin}, \textit{Wheel}, \textit{Bearing}, \textit{Nut}, \textit{Base}, \textit{Axle}, and \textit{Hanger}. These parts are relatively few in number, dimensionally and geometrically diverse, and require various tools and behaviors to assemble. Conveniently, there are numerous online tutorials that describe in layman's terms how to assemble a skateboard truck by hand, to which ChatGPT would have been exposed during training.

We conducted tests on \textit{Gripper Selection}, \textit{Debugging Scripts}, and \textit{Robotic assembly} to answer questions such as (1) What processes does the workflow simplify (2) What's the extent of generalization ChatGPT has to new unseen tool-sets and (3) What limitations does the workflow run into?

\begin{figure}[thpb]
    \centering
    \includegraphics[scale=1.0, width=3.1in]{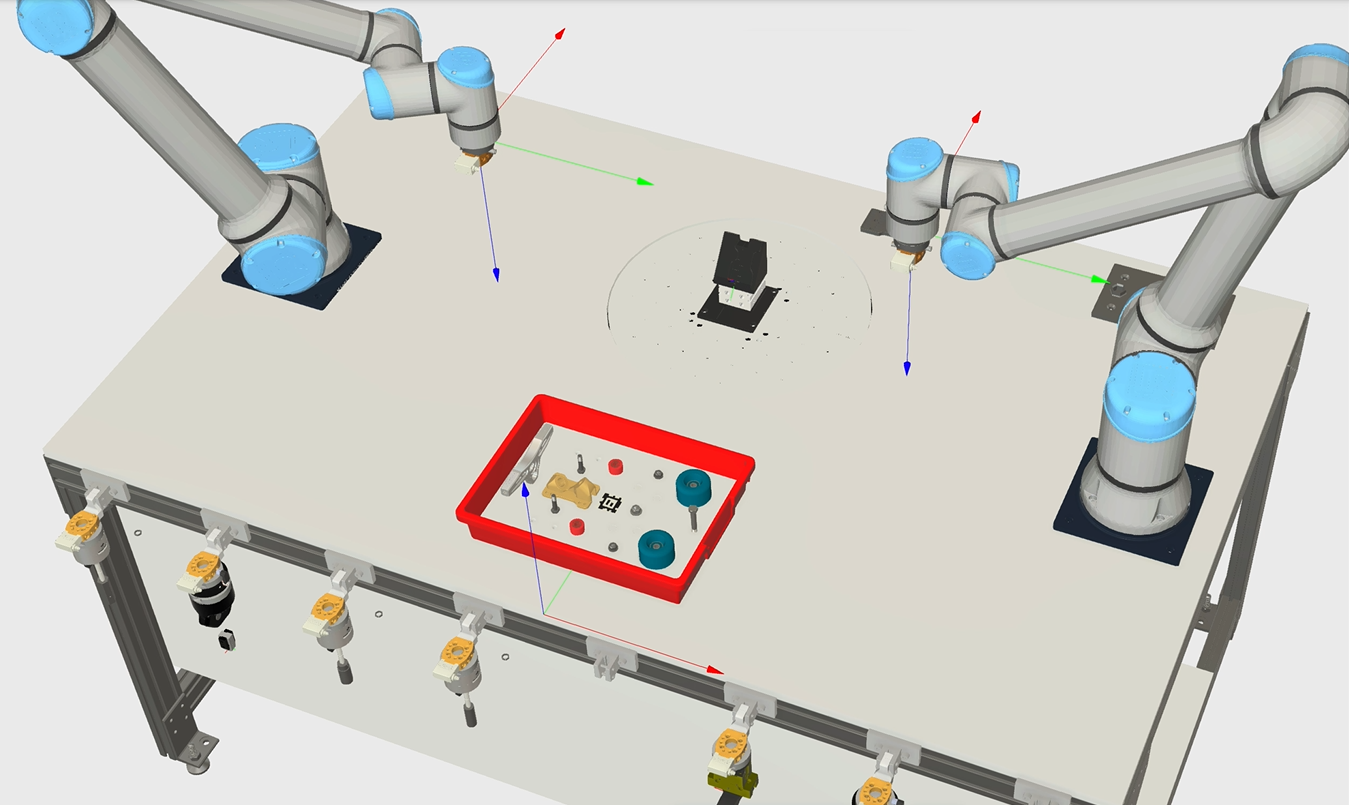}
    \caption{Digital twin of workcell in simulation platform. The workcell contains tool changers hanging off the table, a red kit of parts containing organized skateboard truck pieces, a black vise fixture to hold the skateboard truck pieces and two UR-10e robots.}
    \label{fig:workcell1}
\end{figure}

All experiments are conducted using a closed-source robotics simulation platform integrated with Fusion 360. Our workcell consists of two UR-10e robots mounted to a table and equipped with a gripper and camera. Along one side of the table is a tool rack with alternative grippers that can be interchanged as required by the experiment. Between the robots lies a bin-picking station containing either kitted or assorted parts. Directly opposite is an assembly station containing a vice. An overview of the workcell is shown in Fig. \ref{fig:workcell1}.

\subsection{Gripper Selection}
This experiment evaluates the ability for ChatGPT to select the best tool for picking or fastening a part among a varied selection of grippers. Fastening hardware, such as socket head screws, bolts, or nuts (lock-nuts, wing-nuts etc.), require a high level of dexterity to grasp and manipulate correctly. We simplify the process by utilizing custom grippers to ensure a secure, successful grasp. We provide the SGA a list of the grippers available, API calls to access tools, and a language description detailing the kind of part each gripper is intended to handle or best suited to grasp. The grippers tested include a \textit{Custom Kingpin Gripper}, an \textit{All-Purpose Gripper}, \textit{Ratcheting Grippers}, and a \textit{Custom Baseplate Gripper}. 

\begin{figure}[thpb]
    \centering
    \includegraphics[scale=1.0, width=3.2in]{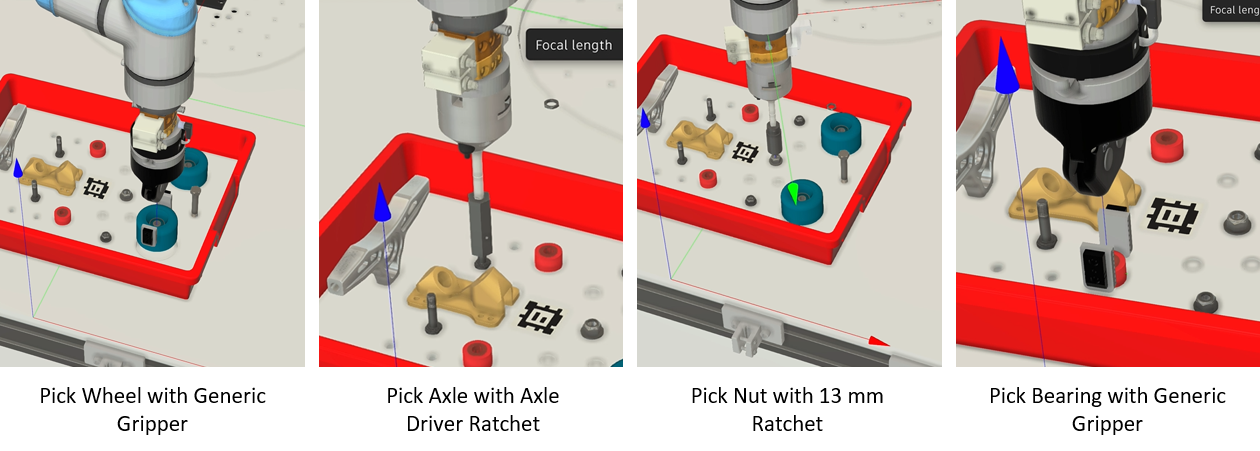}
    \caption{Examples of ChatGPT choosing the best gripper to pick the part.}
    \label{fig:picking-grippers}
\end{figure}

We test between a Generic Language Description (GLD) of each part and a CAD-Derived Language Description (DLD) from the CAD model part-name created by the designer or manufacturer. If the result is incorrect, we send the result and history back through the retry loop requesting a different gripper. The success rates (SR) after three trials are shown in Table \ref{demo-table}. ChatGPT performs well at selecting the correct gripper the first time. In the case it doesn't such as with the kingpin we found a pass through the retry loop successfully fixed this issue. We observe that part-names inherited from CAD models often contain obscure naming conventions which may make it difficult for ChatGPT to understand the functionality of the part. Thus, as touched in \cite{name}, without keywords and descriptive naming conventions in CAD models, this level of generalization would not be achievable. As a result, adopting conventions that store semantic information within a part-name is incredibly useful for LLM based workflows. 

\begin{table}[!h]
\begin{center}
\begin{tabular}{ |p{0.90cm}|p{0.50cm}|p{3.6cm}|p{0.50cm}|p{0.90cm}|  }
\hline
\textbf{GLD} & \textbf{SR\%} & \textbf{DLD} & \textbf{SR\%} & \textbf{SR\% w/retry} \\
\hline
Kingpin & 100 & Kingpin-Bolt-91257A662-Zinc-Plated-Hex-Head-Screw & 0 & 100\\
\hline
Wheel & 100 & Powell-Peralta-90a-art-bones-wheel & 100 & 100  \\
\hline
Bearing & 100 & Hardcore-Bearing & 100 & 100\\
\hline
Nut    & 100 & Kingpin-Nut-93298A135-Medium-Strength-Steel-Nylon-Insert-Flange-Locknut & 100 & 100\\
\hline
Base &   100 & Aera-Baseplate-Pneumatic-Fixture-v26 & 100& 100\\
\hline
Axle & 100 & Aera-Trucks-4140-Axle-+4MM & 100& 100\\
\hline
Hanger & 100 & Area-K4-Hanger & 100 & 100\\
\hline
\end{tabular}
\label{demo-table}
\caption{Gripper Selection Success Rates (SR)}
\end{center}
\end{table}

\subsection{Debugging Scripts}

In this experiment, we explore script generation and debugging to execute a simple motion task: \textit{Move the robot to 100 random positions}. We bootstrap the agent history with an example of such a script, which adds a random float to each component of the robot's current pose. The script, however, is intentionally flawed: (1) an early and unnecessary Exception is raised on purpose (2) the randomness results in unreachable poses, and (3) runtime Exceptions raised by the motion command aren't handled and will cause the script to crash.

The experiment concluded after two iterations. On the first iteration, ChatGPT identified and commented out the Exception in the example script and returned the rest unchanged. When the script was executed, the call to \verb|gripper.move_cartesian| triggered a \verb|MotionException| with the note \textit{unreachable position}, as expected. Provided this exception on the second iteration, ChatGPT reduced the random range by a factor of 10. This is an extremely conservative approach, and the authors would have preferred incremental adjustment and matching edits to the printout "Generating a wild transform". During the same iteration, ChatGPT also incorporated a try/except block to handle future runtime exceptions when moving. Disappointingly, it did not specify the exact exception raised earlier and the introduction of this block, while making the script more likely to finish, means that the robot may not move to all 100 positions as originally requested, due to unreachable positions being skipped. This may indicate a bias in the model towards ensuring code runs without error, even if it may compromise functionality. Following these changes, the script completed without error.

\begin{figure}[thpb]
    \centering
    \includegraphics[scale=1.0, width=3.1in]{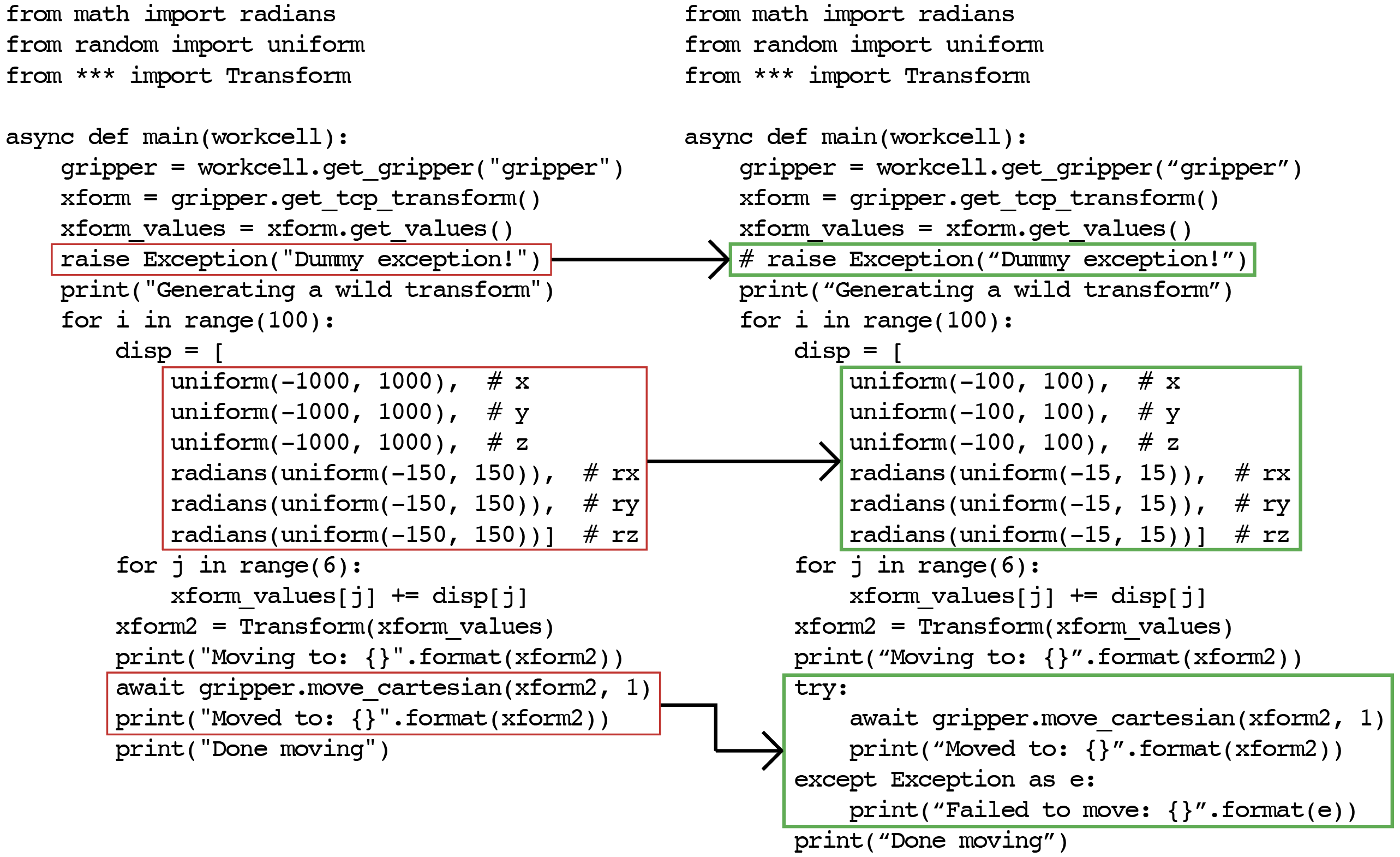}
    \caption{Input (L) and output (R) from SGA for random motion experiment.}
    \label{fig:workcell}
\end{figure}

\subsection{Robotic Assembly}

In this experiment, we explore script generation for an insertion task for one of the the skateboard truck parts after the assembly has been processed by the TDA, namely: \textit{Place Kingpin Bolt on Baseplate}.

The produced script imports the required modules, defines a \verb|main| function with \verb|workcell| as an input parameter, and provides a doc-string describing what the function does and specifies the subtask. It also walks through a series of computational steps to calculate the position and orientation of the kingpin bolt, a somewhat intricate challenge requiring matrix multiplication. It appears to do well amalgamating information in the history, and it accesses the correct constants from the assembly and workcell specifications despite being provided higher-level terms. Before and after placement, the gripper moves to a retracted position, showing consideration for collision avoidance as in the examples. While the SGA relies on templates and examples, its capability to dynamically adjust and produce task-specific code, such as for grasping the unique Kingpin and placing it precisely, is demonstrated in a coherent, functional script. Additionally, it provides numerous comments throughout the script that separate each distinct stage.

\begin{figure}
    \centering
    \includegraphics[scale=1.0, width=3.1in]{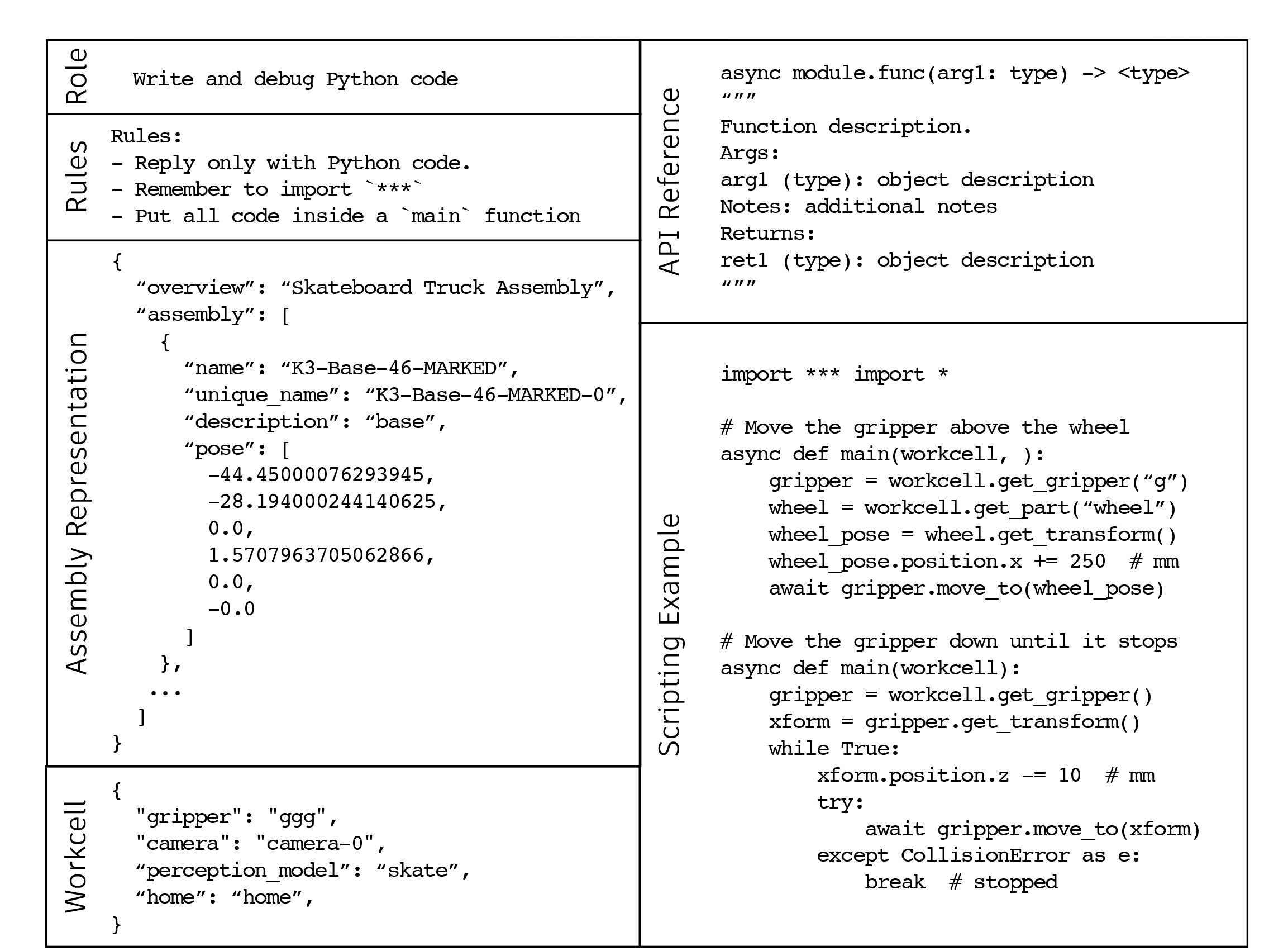}
    \caption{Example of history used to bootstrap a typical SGA on initialization. Note that what's shown is a selection, and that entries are significantly longer in practice.}
    \label{fig:typical-history}
\end{figure}

When executed in simulation after the preceding script, \textit{Pick Kingpin Bolt}, the script ran successfully. Subsequently, \textit{Insert Part on Kingpin Bolt}, also ran successfully. Specifically, the initial and final workcell states for this script were compatible with those of the surrounding scripts, allowing the script to function as a harmonious link in a sequence of operations. This is interesting, as the SGA is unaware of the desired states or actions on either side of the script it's generating. It's possible that this information is subtle but apparent in the output of the TDA and chat history, or if such best-practices were seen by ChatGPT during training. Notable, no error checks or success criteria are implemented, and the script assumes certain prerequisites about the workcell, such as the part being available in the designated location.

\begin{figure}
    \centering
    \includegraphics[scale=1.0, width=3.1in]{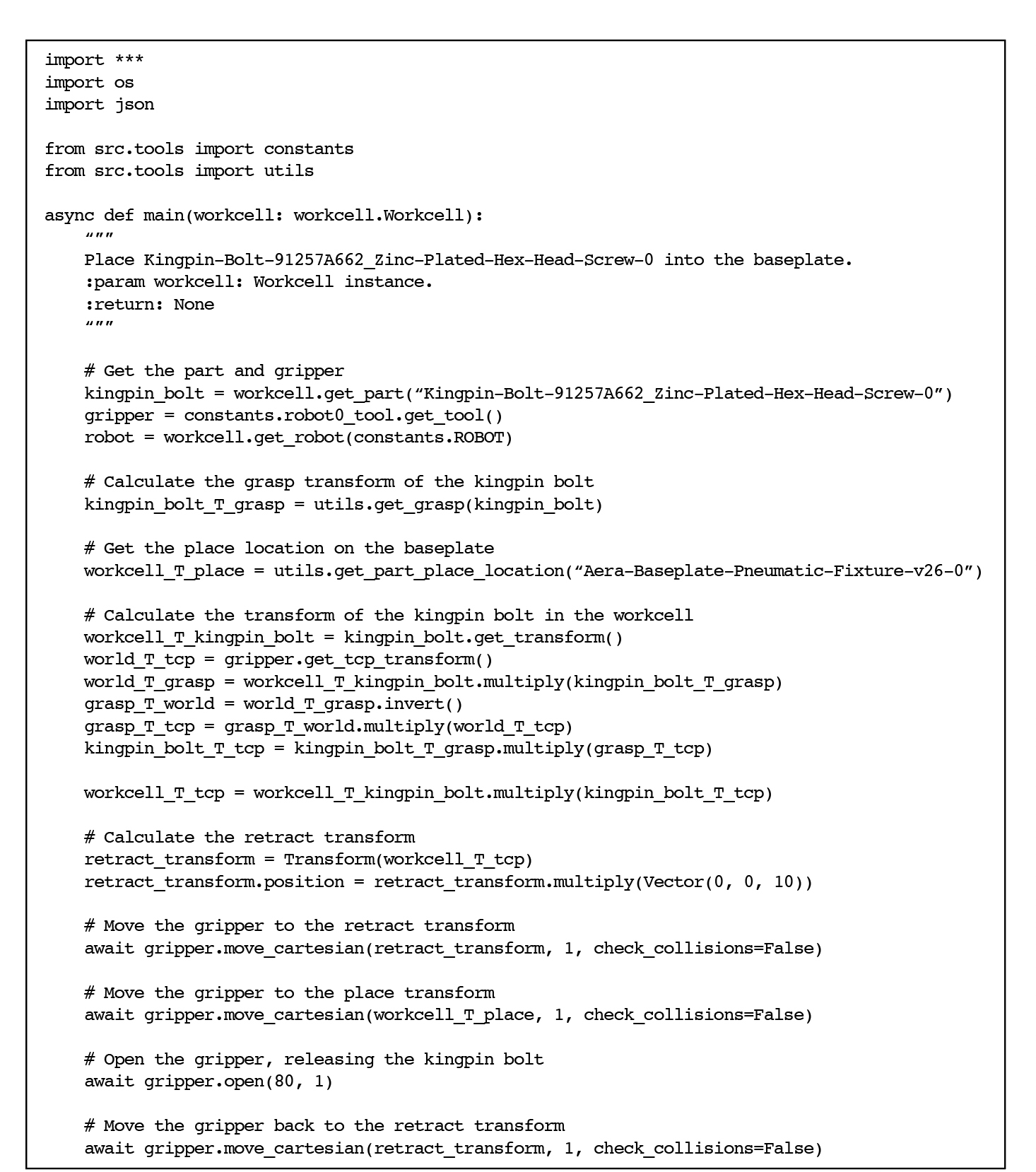}
    \caption{Script generated by the SGA: \textit{Place Kingpin Bolt on Baseplate}}
    \label{fig:typical-history}
\end{figure}

\section{CONCLUSIONS}

This research offers a glimpse into the possibility of using LLMs, like ChatGPT, to automate the coding process for robotic assembly tasks, a process traditionally marked by labor intensiveness and need for expertise. We offer a practical approach to implementing such an automated programming system, and demonstrate its efficacy for basic robotic manufacturing tasks. We recognize the necessity of refining this approach, however, as it has several key limitations.

While the model demonstrated an impressive ability to generate code, our experiments highlight areas where careful human oversight remains crucial, particularly in tasks that demand nuanced understanding of the task and complex spatial reasoning abilities, such as ensuring scripts achieve their intended outcomes and executing dexterous manipulations like re-grasping. These areas, often intuitive for human programmers, show clear gaps in ChatGPT's capabilities for programming successful robotics tasks. By fine-tuning the model on a dataset of prior coding and manufacturing examples, however, there may be an opportunity to not only generalize it for these purposes, but also to reduce the need for the intricate, meticulously configured prompts we employed. While providing robust example code can reduce the likelihood of some errors downstream, real-time debugging of robot programs remains challenging for text-only LLMs, and we're eager to experiment with language models that have innate visual and spatial reasoning skills to overcome these challenges. Looking forward, we're enthusiastic about extending this approach to a broader range of assembly tasks, including those with unique geometries, intricate spatial relationships, and uncommon assembly methods – things that ChatGPT might not have seen during training – pushing the boundaries of what is currently achievable with LLM-driven robot programming.

\addtolength{\textheight}{-12cm}


\end{document}